\documentclass{article}

\PassOptionsToPackage{numbers, sort}{natbib}
\usepackage{natbib}

\usepackage[final]{compositional_learning}

\usepackage[utf8]{inputenc} 
\usepackage[T1]{fontenc}    
\usepackage{hyperref}       
\usepackage{url}            
\usepackage{booktabs}       
\usepackage{amsfonts}       
\usepackage{nicefrac}       
\usepackage{microtype}      
\usepackage[dvipsnames]{xcolor}       

\usepackage{multirow,multicol}
\usepackage{graphics}
\usepackage{graphicx} 
\usepackage{amsmath,amssymb,amsthm}
\usepackage{caption,subcaption}
\usepackage{bbding}
\usepackage{multicol}

\newcommand\blfootnote[1]{%
  \begingroup
  \renewcommand\thefootnote{}\footnote{#1}%
  \addtocounter{footnote}{-1}%
  \endgroup
}

\newcommand{\proposed}{Conditional Autoregressive Slot Attention}
\newcommand{\proposedshort}{CA-SA}
\newcommand{\setsep}{ \ , \ }

\title{Object-Centric Temporal Consistency via Conditional Autoregressive Inductive Biases}

\author{%
  Cristian Meo$^{*,1}$ 
  \And
  Akihiro Nakano$^{*,2}$  
  \AND
  Mircea Lică$^1$
  \And
  Aniket Didolkar$^3$
  \And
  Masahiro Suzuki$^2$
  \And
  Anirudh Goyal$^4$
  \And
  Mengmi Zhang$^{5,6}$
  \And
  Justin Dauwels$^1$
  \And
  Yutaka Matsuo$^2$
  \And
  Yoshua Bengio$^{3}$
}

\begin{document}

\maketitle

\begin{abstract}
 Unsupervised object-centric learning from videos is a promising approach towards learning compositional representations that can be applied to various downstream tasks, such as prediction and reasoning. Recently, it was shown that pretrained Vision Transformers (ViTs) can be useful to learn object-centric representations on real-world video datasets. However, while these approaches succeed at extracting objects from the scenes, the slot-based representations fail to maintain temporal consistency across consecutive frames in a video, i.e. the mapping of objects to slots changes across the video. To address this, we introduce \proposed~(\proposedshort), a framework that enhances the temporal consistency of extracted object-centric representations in video-centric vision tasks. Leveraging an autoregressive prior network to condition representations on previous timesteps and a novel consistency loss function, \proposedshort~predicts future slot representations and imposes consistency across frames. We present qualitative and quantitative results showing that our proposed method outperforms the considered baselines on downstream tasks, such as video prediction and visual question-answering tasks.
\end{abstract}

\section{Introduction}
\vspace{-0.1cm}

The main goal of object-centric (OC) representation learning is to represent each object in an image as a set of separate fixed-size vector representations called ``slots''~\citep{locatello2020object, burgess2019monet, assouel2022objectcentric, dittadi2021generalization, didolkar2023cycle}. This slot-based representation serves to represent natural scenes as a composition of objects~\citep{dittadi2021generalization, locatello2020object, goyal2021recurrent}. Due to this compositional nature of scenes~\citep{greff2020binding}, object-centric representations can enhance out-of-distribution generalization~\citep{dittadi2021generalization}, and handle complex tasks such as reasoning~\citep{assouel2022objectcentric,yang2020object,wu2023slotformer,wu2023slotdiffusion,mansouri2024object}, planning~\citep{pmlr-v100-veerapaneni20a,nakano2023interactionbased},  control~\citep{zadaianchuk2021selfsupervised,mambelli2022compositional,biza2022binding}, and reinforcement learning~\citep{didolkar2023cycle, ferraro2023focus, zadaianchuk2021selfsupervised, yoon2023investigation}. Moreover, object-centric representation learning is in line with studies on the characterization of human perception and reasoning~\citep{kahneman1992reviewing}, making it a very appealing direction in terms of explainability~\citep{Linardatos2020Explainable} as well. \blfootnote{$^*$ Equal contributor. $^1$ Delft University of Technology, NL. $^2$ The University of Tokyo, JP. $^3$ Mila, University of Montreal, CA. $^4$ Google DeepMind, UK.  $^5$ Deep NeuroCognition Lab, CFAR and I2R, Agency for Science, Technology and Research, SG. $^6$ Nanyang Technological University, SG. \\Corresponding author: \texttt{c.meo@tudelft.nl, nakano.akihiro@weblab.t.u-tokyo.ac.jp}}
 
Although recent OC pipelines succeeds at accurately extracting objects from frames in a video \citep{zadaianchuk2023object, wu2023slotdiffusion, kakogeorgiou2023spot}, a persistent problem when applying object-centric models developed for images~\citep{locatello2020object,singh2022illiterate,seitzer2023bridging} to videos is temporal consistency.
Although learning temporal consistent representations has been a central problem for many years~\citep{erhan2010understanding, bengio2012deep, glorot2011domain, duval2023faenet, goyal2021factorizing, goyal2021neural, zadaianchuk2023object}, learning temporal consistent object-centric representations is particularly difficult as the representations are permutation-equivariant. Prior works utilize various architectural biases to achieve temporal consistency. Some approaches have explored employing prior networks to model temporal consistency expicitly~\citep{singh2022simple,kipf2022conditional,wu2023slotformer,wu2023slotdiffusion}. Other models have directly conditioned the slot representations on previous timesteps~\citep{goyal2020object,goyal2021neural,singh2022simple}.
However, we argue that architectural biases may not always be enough to achieve temporal consistency. Another approach is to add an auxiliary loss in the representation space~\citep{zadaianchuk2023object}. Contrary to \citet{zadaianchuk2023object}, we argue that adding such a loss directly on the slots encourages the representations to be too similar between timesteps, which may hinder the model's ability to generalize to longer sequences.

To mitigate the problem of temporal consistency, we propose~\proposed~(\proposedshort), a model-agnostic module that consists of:
(1) An autoregressive network that predicts the initial slot representations of the current timestep from the previous timestep, to condition the current slot extraction on prior timesteps,
and
(2) A temporal consistency loss between the feature-to-slots attention maps of two consecutive frames, to impose the same slot to attend to spatially similar area of the image.
Through ablations, we show that the combination of the two is the key to learning a more temporally consistent representations.
We present qualitative and quantitative evaluations of the proposed approach on the CLEVRER~\citep{Yi2020CLEVRER} and Physion~\citep{bear2021physion} datasets, showing how objects' temporal consistency improve in terms of downstream task performance.

\vspace{-0.2cm}
\section{Related Works}
\vspace{-0.2cm}

The problem of temporal inconsistency has been studied for many years~\citep{bengio2009slow}. Whenever various image processing algorithms are applied as precursors to video processing, certain temporal inconsistencies can be introduced in the consecutive frames of the video. For example, certain noise reduction algorithms may cause flickering due to slight variations in noise patterns of consecutive frames. To deal with such inconsistencies, previous works have introduced various objectives and priors.~\citet{lai2018learning} introduces a perceptual loss to encourage temporal consistency.~\citet{eilertsen2019single}, introduce two regularization terms that force a frame and its affine transformation to have similar representations. A range of approaches also rely on predicting optical flow or motion information for achieving temporal consistency~\citep{lang2012practical, bonneel2015blind, dong2015region, yao2017occlusion}. 
While these works consider general computer-vision problems, the importance of temporal consistency also applies to video-based object-centric models as well.
To ensure temporal consistency various approaches replace the sampling operation, which introduces the permutation equivariance property of slots \cite{locatello2020object}, by conditioning slots on previous ones ~\citep{singh2022simple,kipf2022conditional,goyal2021neural,wu2023slotformer}. In this work, besides introducing a novel architectural bias, we introduce an auxiliary loss which enforces consistency by optimizing for consecutive attention maps to be similar. 

Related works on object discovery and video downstream tasks are summarized in Appendix \ref{appendix:extended_related_works}.

\vspace{-0.2cm}
\section{Method} \label{method}
\vspace{-0.2cm}
When it comes to modelling sequences with an autoregressive model (e.g., RNN~\cite{elman1990finding}, autoregressive transformer~\cite{vaswani2017attention}), ensuring objects-to-slot consistency is necessary to learn meaningful objects dynamics~\citep{goyal2021factorizing, goyal2021recurrent, goyal2021neural}.
In contrast to most existing methods, which either enforce an architectural bias~\citep{singh2022simple, kipf2022conditional,elsayed2022savi++, wu2023slotformer} or a regularizer to enhance temporally consistent object slots~\cite{didolkar2023cycle, zadaianchuk2023object}, our method proposes to use both. \autoref{tab:related_works} shows a comparison of our proposed method with existing approaches which try to mitigate permutation equivariance property of object-centric representations.

\begin{figure}[tb]
    \centering
    \includegraphics[width=0.98\textwidth]{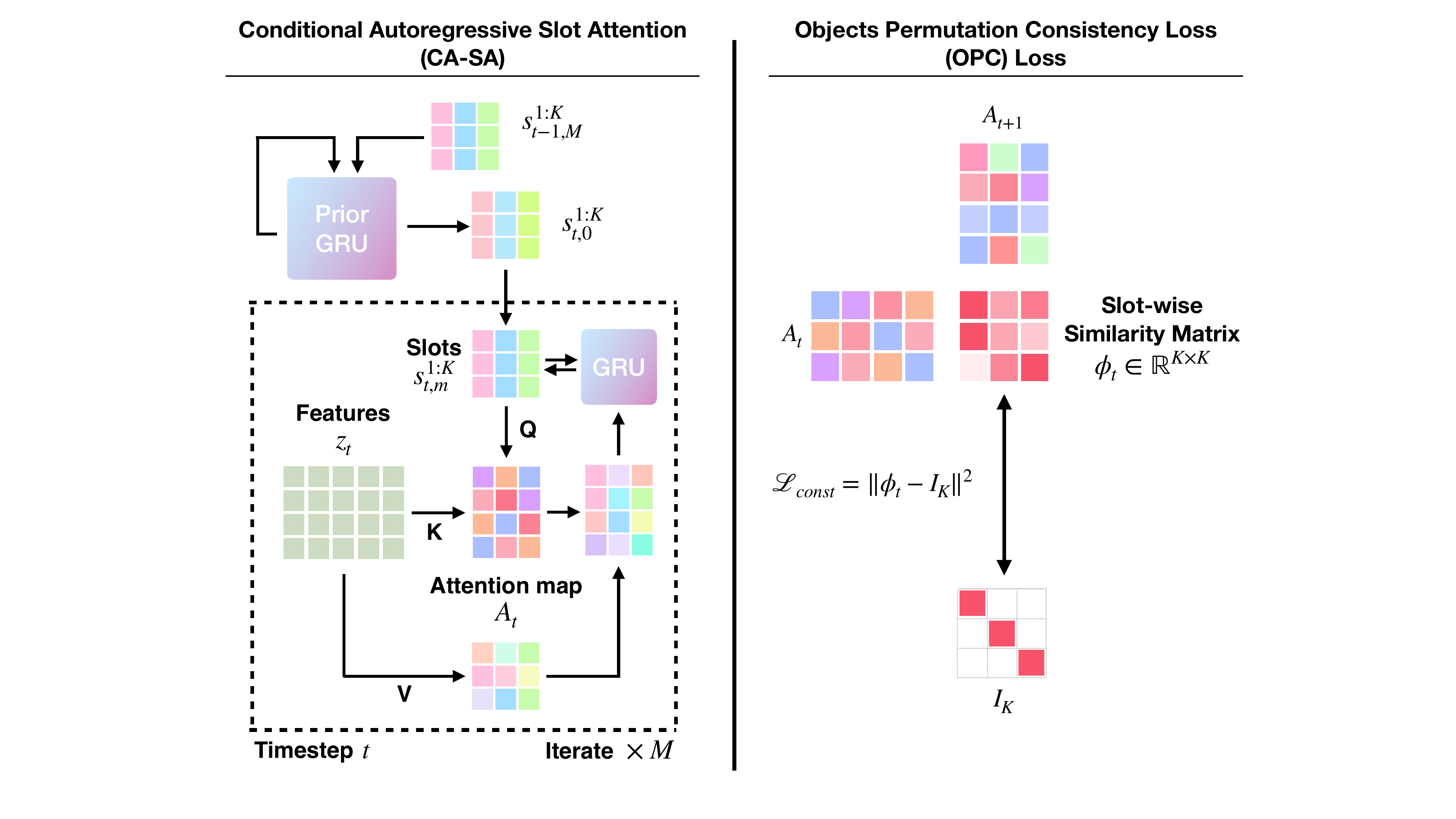}
    \caption{Left: Overall CA-SA architecture is represented. The Prior GRU network takes the slots from the previous timestep and condition the initialization of the new slots. The vanilla SA is represented within the dashed box. Right: Visualization of the OPC loss. Two consecutive attention maps $A_t, A_{t+1}$ are used to compute a cosine similarity distance, whose diagonal elements are optimized to match an identity matrix to impose slots' temporal consistency.}
    \label{fig:architecture}
    \vspace{-0.5cm}
\end{figure}

In this section, we present the two main contributions of our proposed approach, namely, (1) \proposedshort~(\proposed), an autoregressive network that predicts the initial slot representations of the consecutive next timestep and conditions the current slot extraction on prior timesteps, and
(2) OPC (Objects Permutation Consistency Loss), an auxiliary loss between two consecutive attention score matrices of the feature-to-slots attention mechanisms, to impose objects permutation temporal consistency between different timesteps. Our proposed objective and architecture are shown in \autoref{fig:architecture}. 
As our method is architecture-agnostic, this makes it suitable for any SA-based model for videos.

The overall pipeline, frame generation procedure, and preliminaries about Slot Attention~\citep{locatello2020object} can be found in Appendix \ref{appendix:overall_pipeline}.

\subsection{\proposedshort: \proposed}
\paragraph{Conditional Autoregressive Prior.}
Given an input video consisting of $T$ frames $x_{1:T}$, each input image is first individually encoded via a feature extractor to latent features $z_{1:T}$. Then, features are fed into the Slot Attention architecture to infer slot representations $s_{1:T}^{1:K}$. Since our goal is to model an object-centric dynamics of the environment using slot representations, we need to ensure that the same slots are used to represent the same objects in the scene along the whole video trajectory. In this work, we empirically find that updating individual slots via a Gated Recurrent unit (GRU) based prior network yields the best results:
\begin{equation}
    \tilde{s}_{t}^{k}, h_t  = \text{GRU}_{\text{prior}}(s_{t-1}^{k}, h_{t-1}) \setsep k=[1,2,\cdots,K],
\end{equation}
where $s_t,\tilde{s}_{t},h_t$ are the slots, initial slots, and the hidden state of the $\text{GRU}_{\text{prior}}$, respectively. $t$ denotes the timestep. Unlike previous conditioning approaches, which allow for inter-slot interaction using MLP or Transformer, the GRU prior network imposes a structure that prevents representation mixing and preserves the object identity.


\paragraph{OPC: Objects Permutation Consistency Loss.} \label{sec:method_loss}
To define a meaningful consistency loss, we draw inspiration from \citet{spelke2013perceiving}'s findings of how human infants perceive objects using several properties, one of which being their spatiotemporal continuity. 
For Slot Attention, this principle can be translated into the notion that attention maps generated at consecutive timesteps should exhibit consistency, reflecting the assumption that the same object would persist in spatially proximate pixels across successive frames~\cite{chakravarthy2023spotlight}. 
However, when defining such consistency within slots, the imposed inductive bias results to be too strong. 
Indeed, as the loss is backpropagated backward in time through the prior network of slot representations, the cumulative effect of minor alterations in the representations can lead to their deterioration~\citep{nakano2023interactionbased}. 

To solve this issue, we focus on the attention map that is computed within the Slot Attention architecture per timestep. Using attention maps allows us to define a weaker regularization, which does not compromise the slot representation while ensuring slot permutation consistency. Formally, let the attention map at timestep $t$ as $A_t\in\mathbb{R}^{K\times H^\prime W^\prime}$. Given attention maps at consecutive timesteps, $A_t,A_{t+1}$, to encourage the attention maps for the same slot to be consistent over different timesteps, we define OPC as:
\begin{equation}
    \mathcal{L}_{\text{OPC}}=\frac{1}{TK}\sum_{t=1}^T\sum_{i=1}^K\|(\phi_t-I_K)_{ii}\|^2 \setsep \phi_t=\frac{A_tA_{t+1}^{\rm T}}{\|A_t\|\|A_{t+1}\|},
\end{equation}
where $\phi_t$ is the attention-wise cosine similarity between consecutive attention maps and $I_K\in\mathbb{R}^{K\times K}$ is an identity matrix. 
Overall, the proposed method is model-agnostic to any slot-based object-centric learning approach for videos. To incorporate our method, we add the object permutation consistency objective to the original loss function of the method that we wish to apply to. In our case, we optimize the OC feature extractor with a spatial broadcast decoder (SBD)~\citep{locatello2020object, watters2019spatial} to reconstruct the images from slots. The model is trained using an image reconstruction loss as in~\citep{kipf2022conditional} together with our proposed objective: 
\begin{equation}
    \mathcal{L}_{\text{OC-feature extractor}}=\mathcal{L}_{\text{image}}+\lambda\mathcal{L}_{\text{OPC}}
    \setsep \mathcal{L}_{\text{image}} = \text{MSE}(x_{1:T}, \hat{x}_{1:T})
\end{equation}
where $\lambda$ is a hyperparameter. In our experiments, we set the value to $\lambda=0.1$ for all datasets.

\vspace{-0.3cm}
\section{Experiments}
\label{sec:downstream}
\vspace{-0.3cm}
We conduct experiments to evaluate \proposedshort~
by exploring the following question: Do temporally consistent object-centric representations improve their downstream usefulness on video-related tasks?
To answer this question, we validate the proposed model on video prediction (VP) and visual question answering (VQA) using CLEVRER \cite{Yi2020CLEVRER} and Physion \cite{bear2021physion} datasets. We also conduct ablational experiments in \autoref{appendix:ablation}.
We provide further details on the datasets and experimental setup in \autoref{appendix:dataset} and \autoref{appendix:implementation}, respectively.

\begin{figure}[t]
    \caption{Generation results and predicted masks on CLEVRER (above) and Physion (below). Red square indicate slots which temporal consistency is improved by adding \proposedshort.}
    \centering
    \includegraphics[width=\textwidth]{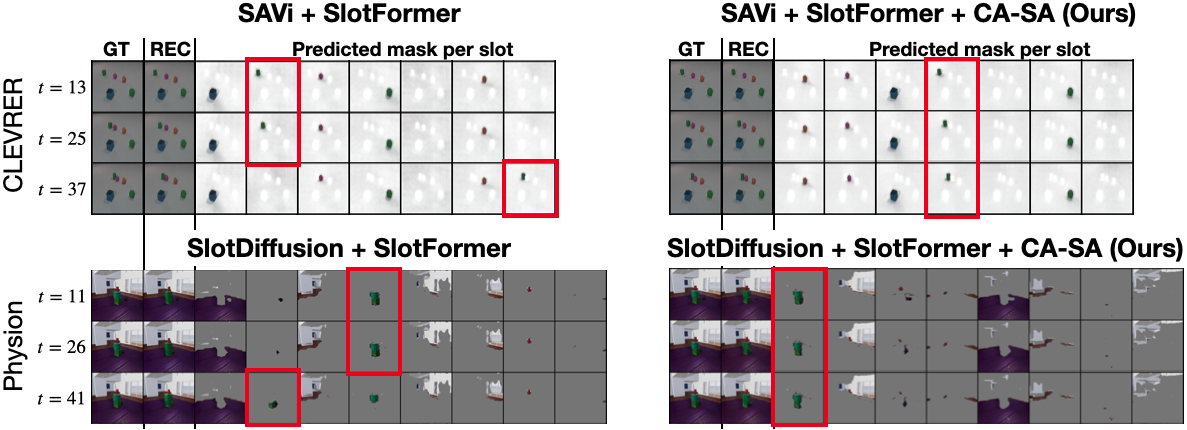}
    \label{fig:rollout_comparison}
    \vspace{-0.3cm}
\end{figure}
\begin{table*}[tb]
  \caption{Evaluation of video prediction task on CLEVRER dataset. $^*$ indicates reproduced results. Best results are indicated in \textbf{bold}.}
  \label{table:vp_clevrer}
  \centering
  \resizebox{\textwidth}{!}{%
  \begin{tabular}{cccc|cccc}
    \toprule
    \multirow{2}{*}{Model} & \multicolumn{3}{c}{Visual quality} & \multicolumn{4}{c}{Object dynamics}\\
     & PSNR ($\uparrow$) & SSIM ($\uparrow$) & LPIPS ($\downarrow$) & AR \% ($\uparrow$) & ARI \% ($\uparrow$) & FG-ARI \%($\uparrow$) & FG-mIoU \% ($\uparrow$) \\
    \midrule
    SAVi-dyn & \textbf{29.77} & \textbf{0.89} & 0.19 & 8.94 & 8.64 & 64.32 & 18.25 \\
    SAVi + SlotFormer$^*$ & 29.22 &  0.87 & 0.15 & 44.19 & 58.49 & 65.96 & 27.90 \\
    \midrule
    \textbf{SAVi + SlotFormer + \proposedshort (Ours)} & 29.47 & 0.88 & \textbf{0.14} & \textbf{46.50} & \textbf{60.52} & \textbf{67.25} & \textbf{28.60}
     \\
    \toprule
  \end{tabular}
  \vspace{-0.8cm}
  }
\end{table*}

\begin{table*}[tb]
  \caption{Evaluation of video prediction task on Physion dataset. $^*$ indicates reproduced results. Best results are indicated in \textbf{bold}.}
  \label{table:vp_physion}
  \centering
  \begin{tabular}{cccc}
    \toprule
    Model & MSE ($\downarrow$) & LPIPS ($\downarrow$) & FVD ($\downarrow$) \\
    \midrule
    STEVE + SlotFormer & 832.0 & 0.43 & 930.6 \\
    SlotDiffusion + SlotFormer$^*$ & \textbf{489.5} & \textbf{0.27} & \textbf{737.8} \\
    \midrule
    \textbf{SlotDiffusion + SlotFormer + \proposedshort (Ours)} & 502.6 & \textbf{0.27} & 759.0 \\
    \toprule
  \end{tabular}
  \vspace{-0.3cm}
\end{table*}

\vspace{-0.3cm}
\subsection{Video Prediction Task} \label{sec:vp}
\vspace{-0.2cm}
\autoref{table:vp_clevrer} and \autoref{table:vp_physion} show the results of the video prediction task for CLEVRER and Physion dataset, respectively. 
~\autoref{fig:rollout_comparison} shows examples of generated slots for both datasets. 
On CLEVRER dataset, as the table shows, \proposedshort~outperforms other baseline models both in terms of visual quality and object-level segmentation. Our model is competitive in terms of visual quality, as the image encoder is the same as in the baseline model. We see that adding temporal consistency improves object-level segmentation for all metrics. 
On Physion dataset, according to \autoref{table:vp_physion} the proposed model performs slightly worse than SlotDiffusion + SlotFormer for MSE and FVD, while tying for LPIPS with a value of $0.27$.

The performance disparity among datasets can be attributed to their distinct characteristics.
Most object-centric models are trained with a surplus of slots compared to the total number of objects in the scene~\cite{locatello2020object, dittadi2021generalization}. As CLEVRER dataset exhibits a simpler background than Physion,
this potentially results in disentanglement with multiple ``empty'' slots, i.e. slots which attend to neither the foreground objects nor the background~\citep{wu2023slotformer}.
Consequently, models trained on CLEVRER show greater performance enhancements over baseline models due to the possibility of temporal inconsistencies arising from empty slots, whereas achieving temporal consistency is more straightforward on Physion.

\begin{table}[t]
  \begin{minipage}{0.49\textwidth}
    \centering
    \caption{VQA task on CLEVRER~\cite{Yi2020CLEVRER}, reporting per-option (per opt.) and per-question (per ques.) accuracy. SF stands for SlotFormer. Both models use Aloe to perform the VQA task. $^*$ indicates reproduced results, best ones are in \textbf{bold}.}
    \label{table:vqa_clevrer}
    \resizebox{0.98\textwidth}{!}{%
    \begin{tabular}{ccc}
      \toprule
      Model & per opt. (\%) & per ques. (\%) \\
      \midrule
      SF + Aloe$^*$ & 90.72 & 80.22 \\
      \midrule
      \textbf{SF + Aloe + \proposedshort~(Ours)} & \textbf{92.69} & \textbf{84.88} \\
      \toprule
    \end{tabular}
    }
  \end{minipage}
  \hfill
  \begin{minipage}{0.49\textwidth}
    \centering
    \caption{VQA task on Physion~\cite{bear2021physion}, reporting accuracy on burn-in (Obs.) and burn-in plus rollout frames (Dyn.). SD and SF stand for SlotDiffusion and SlotFormer, respectively. $^*$ indicates reproduced results, best ones are in \textbf{bold}.}
    \label{table:vqa_physion}
    \resizebox{0.98\textwidth}{!}{%
    \begin{tabular}{ccc}
      \toprule
      Model & Obs. (\%) & Dyn. (\%)\\
      \midrule
      SD + SF$^*$ & 63.8 & 63.9 \\
      \midrule
      \textbf{SD + SF + \proposedshort~(Ours)} & \textbf{64.1} & \textbf{64.7} \\
      \toprule
    \end{tabular}
    }
  \end{minipage}
\end{table}

\vspace{-0.2cm}
\subsection{Video Question Answering Task} \label{sec:vqa}
\vspace{-0.2cm}

\autoref{table:vqa_clevrer} and \autoref{table:vqa_physion} summarize the results on CLEVRER and Physion VQA tasks, respectively. On CLEVER dataset, adding our proposed method improves the VQA accuracy by $1.9\%$ and $4.6\%$ for accuracy per-option and per-question, respectively. On Physion dataset, our model slightly improves accuracy of both metrics. The detailed results of both datasets are in \autoref{appendix:additional_vqa}.
Again, we observe that the performance gain is larger for CLEVRER dataset, as our model is able to reduce the temporal inconsistency caused by empty slots.

\vspace{-0.1cm}
\subsection{Ablation Study} \label{appendix:ablation}
\vspace{-0.2cm}
In this section, we provide ablation results on model architecture of \proposedshort. We report visual quality and object dynamics in video object discovery task of CLEVRER dataset in \autoref{table:ablation}. As the result shows, the combination of using a GRU prior and the proposed auxiliary loss improves over vanilla stochastic SAVi in all metrics except for ARI.

\begin{table*}[t]
  \caption{Ablation study on video object discovery task of CLEVRER dataset.}
  \label{table:ablation}
  \centering
  \resizebox{\textwidth}{!}{%
  \begin{tabular}{cccccc|cccc}
    \toprule
    \multirow{2}{*}{Model} & \multirow{2}{*}{Prior} & \multirow{2}{*}{Aux. Loss} & \multicolumn{3}{c}{Visual quality} & \multicolumn{4}{c}{Object dynamics} \\
    & & & PSNR ($\uparrow$) & SSIM ($\uparrow$) & LPIPS ($\downarrow$) & AR ($\uparrow$) & ARI ($\uparrow$) & FG-ARI ($\uparrow$) & FG-mIoU ($\uparrow$) \\
    \midrule
    \proposedshort & \textcolor{green}{\CheckmarkBold} (GRU) & \textcolor{green}{\CheckmarkBold} & \textbf{40.67} & \textbf{0.98} & \textbf{0.07} & \textbf{78.98} & 79.19 & \textbf{93.94} & \textbf{40.71} \\
    \proposedshort~w/o prior  & \textcolor{red}{\XSolidBrush} & \textcolor{green}{\CheckmarkBold} & 39.32 & 0.97 & 0.08 & 76.28 & 79.15 & 93.83 & 39.54 \\
    \proposedshort~w/o aux. loss & \textcolor{green}{\CheckmarkBold} (GRU) & \textcolor{red}{\XSolidBrush} & 38.92 & 0.97 & 0.08 & 38.12 & 61.28 & 93.60 & 35.32 \\
    \midrule
    StoSAVi & \textcolor{green}{\CheckmarkBold} (MLP) & \textcolor{red}{\XSolidBrush} & 39.81 & 0.97 & 0.08 & 80.47 & \textbf{79.44} & 93.91 & 40.51 \\
    \toprule
  \end{tabular}
  \vspace{-0.6cm}
  }
\end{table*}

\vspace{-0.2cm}
\section{Conclusion}
\vspace{-0.2cm}
In this paper, we proposed \proposedshort, a model-agnostic module consisting an autoregressive network and OPC, an auxiliary loss aimed to improve object-to-slot temporal consistency of video object-centric models. We experimented on two types of downstream tasks, VP and VQA, and showed that adding our proposed method on top of state-of-the-art baselines improve their performances. Particularly, while we observed a marginal improvement in the video prediction task, \proposedshort~enhanced the VQA downstream performance across all metrics. We justified such difference considering that, while the VP task relies on the image space, VQA task uses the extracted slots as input space, clearly showing the importance of having temporal consistent slots. 
As in Slotformer \cite{wu2023slotformer}, we observed that the two-stage training strategy harms the model's performance at the early rollout steps.
Exploring joint training of the base object-centric model and the Transformer dynamics module could potentially benefit the performance of both models. 
We also leave investigation of combining our method with other video object-centric models and applying our method on wider variation of downstream tasks as future works. 

\newpage
\bibliography{references}
\bibliographystyle{abbrvnat}

\newpage
\appendix
\section{Overall Pipeline} 
\label{appendix:overall_pipeline}
\begin{figure}[ht]
    \centering
    \includegraphics[width=0.9\textwidth]{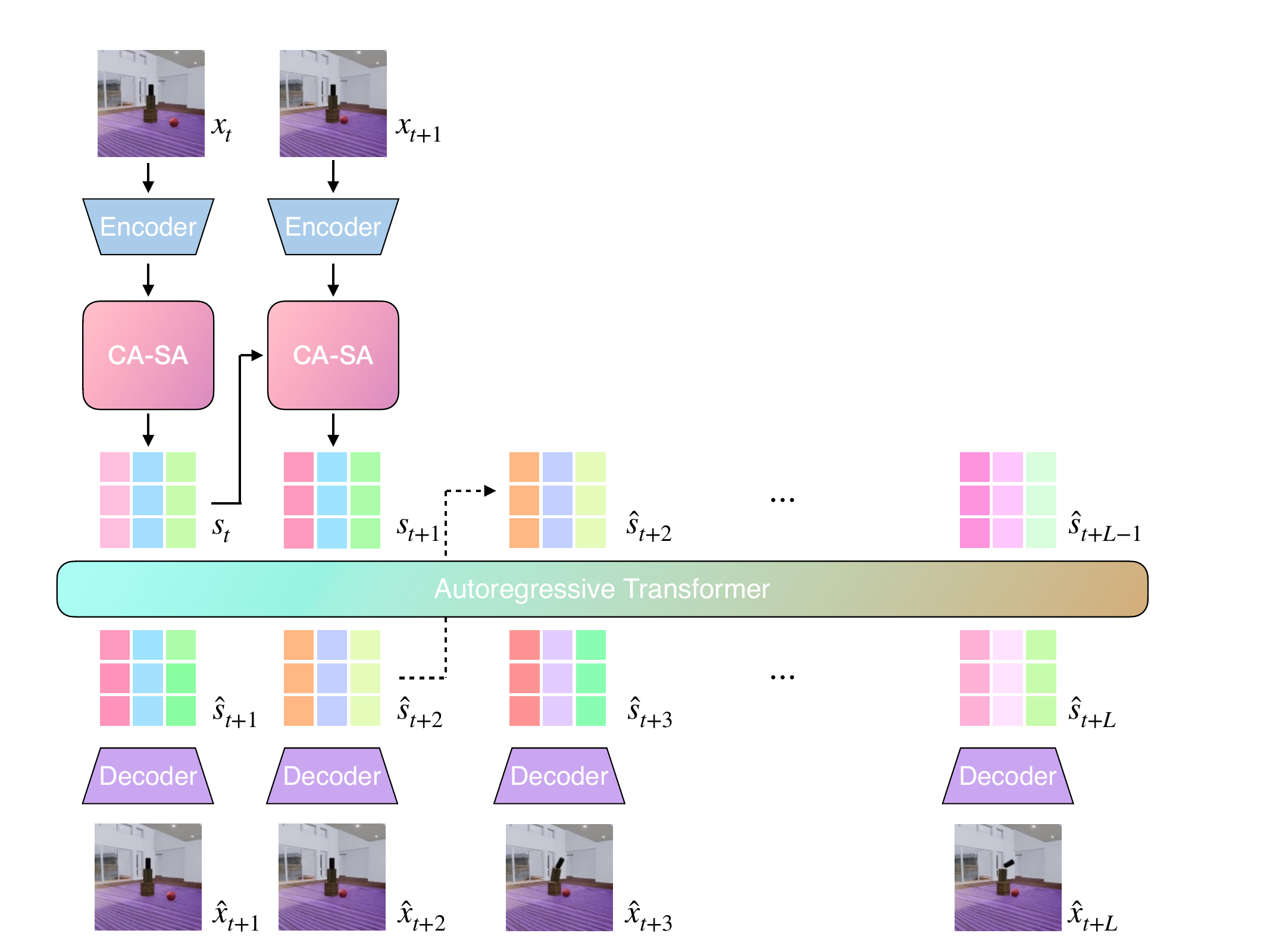}
    \caption{Proposed pipeline: Images $x_t$ are first encoded into features, which are used to extract slots $s_t$. Slots video trajectory is generated using an autoregressive transformer and decoded into the predicted video using a Spatial Broadcast Decoder.}
    \label{fig:architecture-video}
\end{figure}

This section presents the overall pipeline, slot attention preliminaries, and frame generation procedure. \autoref{fig:architecture-video} shows the overall pipeline of \proposedshort, where \proposedshort~is used to extract temporal consistent slots from the frames, and the autoregressive transformer is used to generate future slots. 
\subsection{Preliminaries}
\paragraph{Slot Attention (SA).}
Slot Attention is an architecture proposed for unsupervised object-centric representation learning from images. An input image $x\in\mathbb{R}^{3\times H\times W}$ is processed through a Convolutional Neural Network (CNN) encoder (feature extractor) to extract features $z\in\mathbb{R}^{{D_{enc}}\times H^\prime\times W^\prime}$. Here, $D_{enc}$ is the feature dimension, and $H,W$ and $H^\prime,W^\prime$ are the height and width of the input and encoded image, respectively. The features are then combined with positional embeddings and flattened spatially. Then, the model initializes $K$, $D_{slot}$-dimensional object-centric representations, $\tilde{s}^{1:K}\in\mathbb{R}^{K\times {D_{slot}}}$, from some distribution. 
Using the slots as query and the features as key and value, Scaled Dot-Product Attention~\citep{vaswani2017attention} is calculated for $M$ iterations to update slot representations, $s^{1:K}=f_{SA}(\tilde{s}^{1:K},M)$, where $f_{SA}$ is the SA function. Unsupervised scene decomposition into individual objects is encouraged through the calculation of iterative self-attention by motivating the slots compete against each other to attend to different parts of the image. The slots are then fed to a spatial broadcast decoder (SBD)~\citep{watters2019spatial} to reconstruct the input image. The entire architecture is trained using image reconstruction loss only.

\subsection{Frame Generation using Slot Representations} \label{sec:method_generation}
Given slot representations $s_{1:T}^{1:K}$, we wish to generate a sequence of future slots of length $L$, $\hat{s}_{T+1:T+L}^{1:K}$. To do so, we follow the approach proposed by SlotFormer~\citep{wu2023slotformer} and employ an autoregressive transformer $\mathcal{T}$~\citep{vaswani2017attention} architecture to perform sequence modelling of the extracted slots. 
The autoregressive transformer  input space is defined using a Multi-Layer Perceptron (MLP) layer, ${\rm MLP}_{in}$ which maps slots to embeddings, and positional encodings that are summed to the transformer embeddings to impose a temporal structure. A MLP head ${\rm MLP}_{out}$ is used to map the transformer outputs back to the slot space. Overall, the sequence modelling can be formally expressed as,
\begin{equation}
    u_{1:T}^{1:K}={\rm MLP}_{in}(s_{1:T}^{1:K}) \setsep
    v_{1:T}^{1:K}=\mathcal{T}(\tilde{u}_{1:T}^{1:K}) \setsep
    \hat{s}_{2:T+1}^{1:K}={\rm MLP}_{out}(v_{1:T}^{1:K}),
\end{equation}
where $\tilde{u}_{1:T}^{1:K} = u_{1:T}^{1:K} + p_t$, $p_t$ are the positional encodings~\citep{vaswani2017attention} and each slot representation is used to predict the same slot at the next timestep. To generate a full trajectory, each slot is generated autoregressively following the approach defined by~\citet{wu2023slotformer}. The autoregressive transformer $\mathcal{T}$ is optimized using the following autoregressive objective:
\begin{equation}
    \mathcal{L}_{\text{Dyn}} = \text{MSE}(s_{2:T}^{1:K}, \hat{s}_{1:T-1}^{1:K})+\text{MSE}(x_{2:T}^{1:K}, \hat{x}_{1:T-1}^{1:K}),
\end{equation}
We use the SBD that was trained in \autoref{sec:method_loss} to decode the predicted slots to image space. While training the autoregressive transformer, the weights of the SBD are kept frozen. 



\section{Extended Related Works} 
\label{appendix:extended_related_works}
Object-centric learning has been gathering attention as a promising direction towards learning efficient and compositional representations of complex scenes without supervision~\citep{locatello2020object,greff2019multi,singh2022illiterate,wu2024neural,jia2023improving,didolkar2023cycle}. While recent works have succeeded in applying this approach to real-world scenes~\citep{seitzer2023bridging, zadaianchuk2023object, wu2023slotdiffusion, kakogeorgiou2023spot}, their evaluation is limited to mask-based metrics. Contrary to this, this work focuses on evaluating the quality of the object-centric representations themselves by applying the representations to downstream tasks. Specifically, we focus on two types of downstream tasks - video prediction and visual question answering. While relatively few, there have been some works that have tackled problems similar to the ones we consider here \citep{kipf2019contrastive, goyal2021neural, kesystematic, pmlr-v100-veerapaneni20a, wu2023slotformer,wu2023slotdiffusion}. All of these works, except ~\citep{wu2023slotformer,wu2023slotdiffusion}, employ a factored representation coupled with a recurrent dynamics module for video prediction or world modelling. ~\citet{wu2023slotformer} and ~\citet{wu2023slotdiffusion} adopt a transformer-based dynamics module.
Out of this, ~\citep{goyal2021neural,wu2023slotformer,wu2023slotdiffusion} consider Slot Attention~\citep{locatello2020object} as the base model to extract slots from the model. These models rely on architectural priors to impose temporally consistency between slots from neighbouring timesteps.
Contrary to these methods, our approach introduces an objective that explicitly optimizes for temporal consistency. Moreover, our approach can be integrated into any of the above three approaches. \autoref{tab:related_works} further highlights the differences between the proposed and existing approaches. 

\begin{table}[tb]
    \centering
    \caption{This table serves to highlight the differences of our models with prior works. The column Temporal-Consistency Prior indicates the approach taken by each model to ensure temporally consistent slot representations across frames in a video.}
    \label{tab:related_works}
    \resizebox{\textwidth}{!}{%
    \begin{tabular}{c|ccc|ccc}
    \toprule
         \multirow{2}{*}{Method} & \multicolumn{3}{c}{Temporal-Consistency Prior} & \multicolumn{3}{c}{Tasks} \\
         & Conditioning of slots & Auxiliary loss & Use RGB inputs only & Reconstruction & Prediction & VQA \\
         \midrule
         SCOFF~\cite{goyal2020object} & previous slots & \textcolor{red}{\XSolidBrush} & \textcolor{green}{\CheckmarkBold} & \textcolor{green}{\CheckmarkBold} & \textcolor{green}{\CheckmarkBold} & \textcolor{green}{\CheckmarkBold} \\
         NPS~\cite{goyal2021neural} & previous slots  & \textcolor{red}{\XSolidBrush} & \textcolor{green}{\CheckmarkBold} & \textcolor{green}{\CheckmarkBold} & \textcolor{green}{\CheckmarkBold} & \textcolor{red}{\XSolidBrush} \\
         STEVE~\citep{singh2022simple} & previous slots & \textcolor{red}{\XSolidBrush} & \textcolor{green}{\CheckmarkBold} & \textcolor{green}{\CheckmarkBold} & \textcolor{green}{\CheckmarkBold} & \textcolor{red}{\XSolidBrush} \\
         SAVi~\citep{kipf2022conditional} & Transformer prior & \textcolor{red}{\XSolidBrush} & \textcolor{red}{\XSolidBrush} & \textcolor{green}{\CheckmarkBold} & \textcolor{green}{\CheckmarkBold} & \textcolor{red}{\XSolidBrush} \\
         VideoSAUR~\citep{zadaianchuk2023object} & \textcolor{red}{\XSolidBrush} & \textcolor{green}{\CheckmarkBold} & \textcolor{green}{\CheckmarkBold} & \textcolor{green}{\CheckmarkBold} & \textcolor{red}{\XSolidBrush} & \textcolor{red}{\XSolidBrush} \\
         SlotFormer~\citep{wu2023slotformer} & MLP/Transformer prior  & \textcolor{red}{\XSolidBrush} & \textcolor{green}{\CheckmarkBold} & \textcolor{green}{\CheckmarkBold} & \textcolor{green}{\CheckmarkBold} & \textcolor{green}{\CheckmarkBold} \\
         SlotDiffusion~\citep{wu2023slotdiffusion} & Transformer prior & \textcolor{red}{\XSolidBrush} & \textcolor{green}{\CheckmarkBold} & \textcolor{green}{\CheckmarkBold} & \textcolor{green}{\CheckmarkBold} & \textcolor{green}{\CheckmarkBold} \\
         \midrule
         \textbf{Ours} & GRU prior & \textcolor{green}{\CheckmarkBold} & \textcolor{green}{\CheckmarkBold} & \textcolor{green}{\CheckmarkBold} & \textcolor{green}{\CheckmarkBold} & \textcolor{green}{\CheckmarkBold} \\
         \toprule
    \end{tabular}
    }
    \vspace{-0.3cm}
\end{table}

\section{Dataset Details} \label{appendix:dataset}
We validate the proposed model on the Video Prediction and Visual Question Answering downstream tasks on CLEVRER and Physion datasets. In this section, we provide further details on the dataset and preprocessing of the data. 

\paragraph{CLEVRER.}
CLEVRER \cite{Yi2020CLEVRER} consists of realistically rendered sequences with multiple 3D objects moving in the scene. The objects differ in shape, color, and texture. The size of each object are kept identical so that no vertical bouncing occurs during collision. The dataset, similar to CLEVR~\citep{johnson2017clevr} and OBJ3D~\citep{lin2020improving}, features smaller objects and more diverse interactions of objects, making it a more challenging task. The attributes of the objects are randomly sampled under the constraint that none of the objects in the scene have the identical attributes. Objects' positions are randomly initialized for each sequence. For each sequence, some objects are randomly chosen such that they cause a collision with each other. The dataset is accompanied by a VQA task with four types of questions: descriptive, explanatory, predictive, and counterfactual. Descriptive questions focus on understanding the video's dynamic content and temporal relations, asking about objects' attributes in an open-ended format. Explanatory questions explore causal relationships, asking which objects or events are responsible for other events. Predictive questions test the ability to predict future events. Counterfactual questions evaluate the understanding of hypothetical scenarios by asking what would or would not happen under altered conditions. Descriptive questions are open-ended questions, while the other three questions are in multiple-choice format with more than one possible answer. 

\paragraph{Physion.}
Physion \citep{bear2021physion} consists of eight video categories, each showing a different physical phenomenon, such as rigid- and soft-body collisions, falling, rolling, and sliding motions. Each video category presents foreground objects, which vary in categories, textures, colors, and sizes, and diverse background scenes environment showed from randomized camera poses. 

The Physion dataset consists of three set: Training, Readout Fitting, and Testing. 
Following SlotDiffusion~\citep{wu2023slotdiffusion}, we sub-sample the frames by a factor of 3 for training the dynamics module and truncated by 150 frames, since that is the threshold within most of interactions happen. To validate models performances we adopt the official evaluation protocol. First, the dynamics models are trained on videos from the Training set. Then, conditioned by the first 45 frames of Readout Fitting and Test videos, they perform rollout to generate future scene representations. A linear readout model is trained on observed and rollout scene representations from the Readout Fitting set to classify whether an ``agent'' object (colored in red) contact with the ``patient'' object (colored in yellow) as the scene unfolds. The classification accuracy of the trained readout model on the Testing set scene representations is reported. For detailed descriptions of the VQA evaluation, refer to their paper \citep{bear2021physion}.

\section{Implementation Details} \label{appendix:implementation}
\subsection{Baselines}
We build our model on SlotFormer~\citep{wu2023slotformer} and SlotDiffusion~\citep{wu2023slotdiffusion} for CLEVRER and Physion dataset, respectively. Their implementations are available online.\footnote{\url{https://github.com/pairlab/SlotFormer}}\footnote{\url{https://github.com/Wuziyi616/SlotDiffusion}}

\paragraph{Stochastic SAVi (StoSAVi).}
As described by \citet{wu2023slotformer}, vanilla SAVi~\citep{kipf2022conditional} occasionally fails to capture objects newly entering the scene.~\citet{wu2023slotformer} explains that this is caused by the more than one ``empty'' slots competing against each other to attend to the newly entered object, resulting in multiple slots representing the same object. To solve this problem,~\citet{wu2023slotformer} proposes a stochastic version of SAVi, in which slots are initialized conditioned on previous timesteps added with a sampling procedure.

Specifically, the output of the prior network is processed through a two-layer MLP with Layer Normalization~\citep{ba2016layer} to predict the mean and log variance of the initial slots at the next timestep:
\begin{equation}
    \tilde{s}_{t}^{k}\sim \mathcal{N}(\mu_t^k, \{\log\sigma_t^2\}^k) \setsep 
    (\mu_t^k, \{\log\sigma_t^2\}^k) = \text{MLP}(f_{\text{prior}}(s_{t-1}^{k}))
\end{equation}
where $f_{\text{prior}}$ is some network used to condition slots on previous timesteps.

The model is optimized by adding a KL divergence loss on the predicted distribution to the image reconstruction loss. The loss only penalizes the log variance with a prior value $\hat{\sigma}$:
\begin{equation}
    \mathcal{L}_{\text{KL}}=\frac{1}{TK}\sum_{t=1}^T\sum_{k=1}^K 
    D_{\text{KL}}(\mathcal{N}(\mu_t^k, \{\log\sigma_t^2\}^k) \| \mathcal{N}(\mu_t^k, \{\log\hat{\sigma}^2\}^k))
\end{equation}
We set $\hat{\sigma}=0.1$ for all datasets. The coefficient of this loss is set to $1\times 10^{-4}$. We follow the same model architecture as implemented in ~\citep{wu2023slotformer}.

\paragraph{SlotDiffusion.}
The model is trained in two-stage manner, by first pretraining a VQVAE~\citep{van2017neural} to convert images to tokenized patches, and then train the encoder and Slot Attention architecture. We follow the same model architecture and training settings as ~\citep{wu2023slotdiffusion}, where the encoder is a modified ResNet18 encoder~\citep{kipf2022conditional} and the decoder is LDM-based~\citep{rombach2022high} trained to predict the noise $\epsilon$ added to the features $z$ obtained by the pretrained VQVAE.

\paragraph{SlotFormer.} 
After training an arbitrary object-centric model, the slots are extracted for all videos and saved offline. Then, SlotFormer is trained to predict slots at future timesteps, conditioned on burnin frames. The architecture and training strategy are kept unchanged from~\citep{wu2023slotformer}.

\subsection{Proposed Approach: \proposedshort}
We implement our prior network using a GRU network. As we implement our method on top of StoSAVi, the initial slots at timestep $t$ are sampled using the predicted mean and log variance which are computed as,
\begin{equation}
    \tilde{s}_{t}^{k}\sim \mathcal{N}(\mu_t^k, \{\log\sigma_t^2\}^k) \setsep 
    (\mu_t^k, \{\log\sigma_t^2\}^k) = \text{MLP}(\text{GRU}_{\text{prior}}(s_{t-1}^{k})).
\end{equation}
We omit the hidden states $h_t$ for simplicity. Following StoSAVi, the KL divergence loss is added to the total loss.

As described in \autoref{sec:method_loss}, the consistency loss is calculated per slot at each timestep and averaged over them. The coefficient of the loss term is set to $\lambda=0.1$.
To use image reconstruction loss when training the autoregressive transformer and to visualize the predicted slots, we train a CNN-based spatial broadcast decoder separately. This decoder is trained using reconstruction loss in image space, and the loss is not backpropagated to the encoder.

We follow Slotformer \cite{wu2023slotformer} approach to evaluate VP and VQA downstream tasks. Specifically, we first train \proposedshort, then train the autoregressive Transformer as described in \autoref{sec:method_generation} using the inferred slots from the model. We validate both downstream tasks on CLEVRER \cite{Yi2020CLEVRER} and Physion \cite{bear2021physion} datasets.

For CLEVRER dataset, we apply \proposedshort~on top of SlotFormer~\citep{wu2023slotformer}, while for Physion we use SlotDiffusion~\citep{wu2023slotdiffusion} as backbone model, as they are the state-of-the-art models on respective datasets. 
To have a fair comparison we adopt the spatial broadcast decoder used by Slotformer \cite{wu2023slotformer} and the conditional latent diffusion model used by SlotDiffusion \cite{wu2023slotdiffusion}, respectively. 

To perform the VQA task, we train an auxiliary model using the slot representations generated by the autoregressive Transformer as inputs. On CLEVRER VQA task, we employ Aloe~\citep{ding2021attention}, a Transformer-based architecture that uses slot representations from input frames and text tokens of the question to predict the answer. For predictive questions, we use the trained Transformer to predict slots at future timesteps, and feed them to Aloe. For other questions, we follow the implementation of Aloe.
On Physion VQA task, we follow the official protocol by training a readout model on generated slots, as there is no language involved in the task. Following \citep{wu2023slotformer}, we implement a readout model which consists of a MLP applied on every two slots to extract relations between slots and a max-pool operation which is invariant to input permutations.

On CLEVRER, the training of \proposedshort~using CNN encoder takes 8 hours to train on 4 V100 GPUs. The training of the autoregressive transformer takes approximately 2 days with the same GPU setup. The training of VQA model takes 3 hours. On Physion, the initial training of VQVAE takes 20 hours. The training of SlotDiffusion requires 30 hours of training on 8 A100 GPUs. The training of the autoregressive transformer takes approximately 15 hours on 4 V100s. The training of the readout model finishes in less than 5 minutes.


\begin{table*}[h]
  \caption{Hyperparameters used to train different encoders on each dataset.}
  \label{table:hyperparams}
  \centering
  \begin{tabular}{cccccc|c}
    \toprule
    \textbf{Dataset} & CLEVRER & Physion \\
    \midrule
    Image encoder   & ResNet18  & ResNet18 \\
    Image resolution $(H, W)$ & $(64, 64)$ & $(128, 128)$ \\  
    Length of sequence $T$      & 6 & 3 \\
    \# of features $H^\prime W^\prime$ & 4096 & 1024 \\
    Feature dimension $D_{enc}$ & 128 & 192 \\
    \# of slots $K$         & 7 & 8 \\
    \# of slot attention iteration $M$ & 3 & 2 \\
    Slot dimension $D_{slot}$   & 128 & 192 \\
    Batch size      & 64 & 48 \\
    Training epochs & 12 & 10 \\
    \toprule
  \end{tabular}
\end{table*}
\begin{table}[t]
  \centering
  \caption{Hyperparameters used to train autoregressive transformer on each dataset.}
  \label{table:hyperparams_slotformer}
    \resizebox{0.5\textwidth}{!}{%
    \begin{tabular}{ccc}
      \toprule
    \textbf{Dataset} & CLEVRER & Physion \\
    \midrule
    Burnin frames $T$   & 6 & 10 \\
    Rollout frames $L$  & 15 & 10 \\
    Batch size          & 64 & 128 \\
    Training epochs     & 80 & 25 \\
    \# of layers        & 4 & 12 \\
    \# of heads         & 8 & 8 \\
    Dimension           & 256 & 256\\
    FFN dimension       & 1024 & 1024 \\
    \toprule
    \end{tabular}
    }
\end{table}

\autoref{table:hyperparams} and \autoref{table:hyperparams_slotformer} describes the hyperparameters used in our the experiments.

\subsection{Experimental Setup}
\paragraph{Video Prediction Task.} We compare \proposedshort~with three state-of-the-art, OC models, SAVi-dyn, SAVi + SlotFormer, and SlotDiffusion + SlotFormer. SAVi-dyn uses SAVi~\citep{kipf2022conditional} as the encoder and combines with a Transformer-LSTM to generate future slots. SAVi + SlotFormer and SlotDiffusion + SlotFormer combine respective models. For CLEVRER, the stochastic verison of SAVi was used in order to accomodate to new objects entering the scene during rollout.

\vspace{-0.2cm}
For CLEVRER, we use PSNR, SSIM, and LPIPS to evaluate the visual quality of the frames generated by each model, and ARI, FG-ARI, FG-mIoU, and AR for evaluation of object-level segmentation quality. For Physion, following~\citep{wu2023slotdiffusion}, we report visual quality metrics only, MSE, LPIPS, and FVD~\citep{unterthiner2019fvd}. 

We follow \cite{wu2023slotformer,wu2023slotdiffusion} with the evaluation protocol for both datasets. On CLEVRER, we use 6 burn-in timesteps to condition the model and then perform a rollout to predict the next slots for 10 steps. On Physion, the model was trained using 15 burn-in and 10 rollout timesteps. The predicted slots were decoded to images using the SBD and compared with the ground truth ones.

\vspace{-0.1cm}
\paragraph{Video Question Answering Task.}
For both datasets, we apply \proposedshort~on top of their respective state-of-the-art model. On CLEVRER, we compare against SlotFormer + Aloe (denoted as SF + Aloe)~\citep{wu2023slotformer}. SF + Aloe first trains StoSAVi as the feature extractor, followed by SlotFormer. Then, the predicted slots from SlotFormer and text tokens of the question are used to train Aloe, a Transformer-based VQA model. For Physion, we select SlotDiffusion + SlotFormer as the baseline model (SD + SF)~\citep{wu2023slotdiffusion}. This model first trains SlotDiffusion as the feature extractor, followed by SlotFormer, and finally a readout model using the predicted slots.

We report two types of average accuracy on CLEVRER VQA task, per-option (per opt.) and per-question (per ques.), as the VQA task includes multiple choice questions with more than one possible answers. The per option accuracy assesses the model's overall correctness in selecting individual options across all questions. Conversely, the per question accuracy measures correctness on a question-by-question basis, necessitating the accurate selection of all answer choices for each question. For Physion VQA task, we report the accuracy when using only burn-in frames (denoted as Obs.) and using burn-in frames and rollout frames (Dyn.). 

We follow the implementation of \citet{wu2023slotformer} for evaluation on both datasets. On CLEVRER, we train Aloe~\citep{ding2021attention} using the predicted slots by SlotFormer, generated by the procedure described in \autoref{sec:method_generation}. The slots are concatenated with the text tokens of the questions and then fed to Aloe. On Physion, we train a readout model which receives every two predicted slots at each timestep as inputs. The outputs of the readout model are max-pooled over all pairs of slots and time to predict the answer.



\section{Rollout Visualizations} \label{appendix:rollout_visualization}
\begin{figure}[tb]
    \centering
    \includegraphics[width=\textwidth]{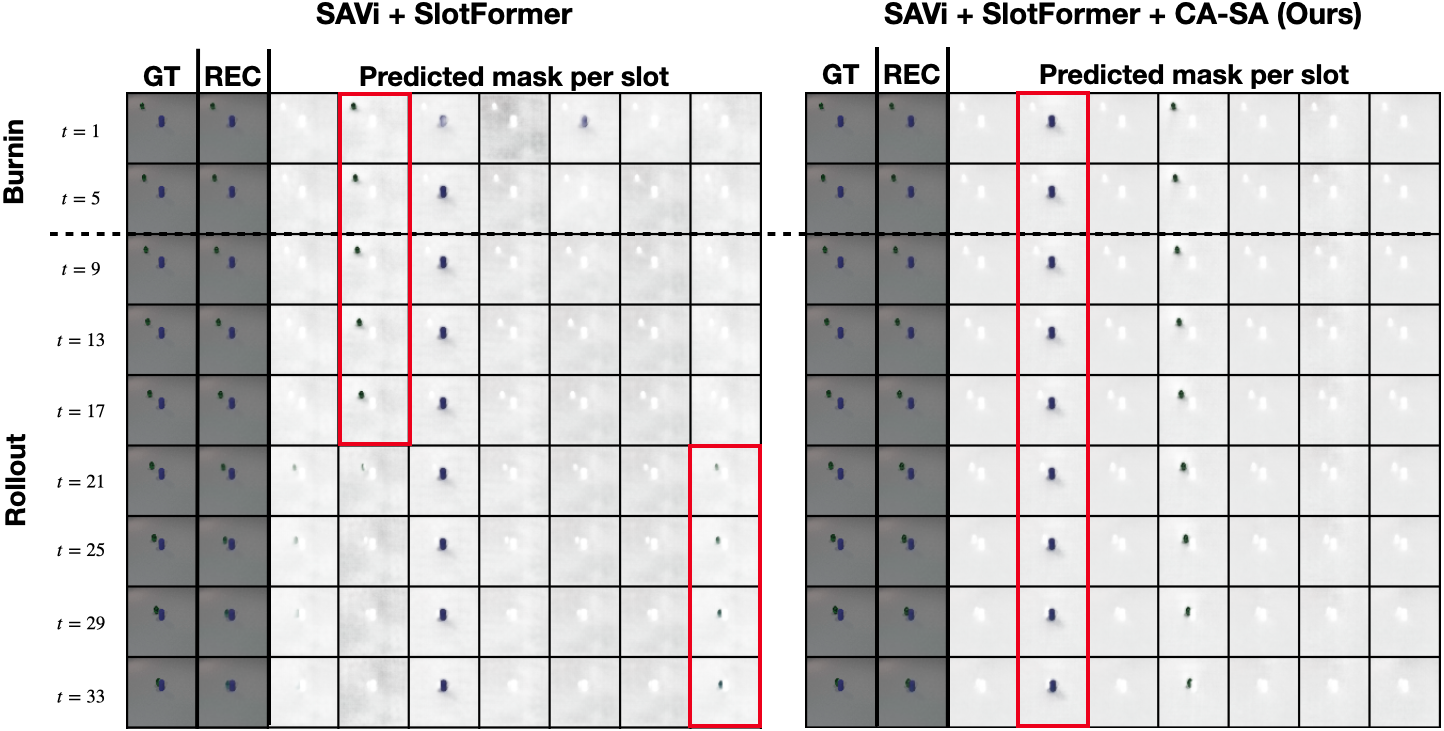}
    \includegraphics[width=\textwidth]{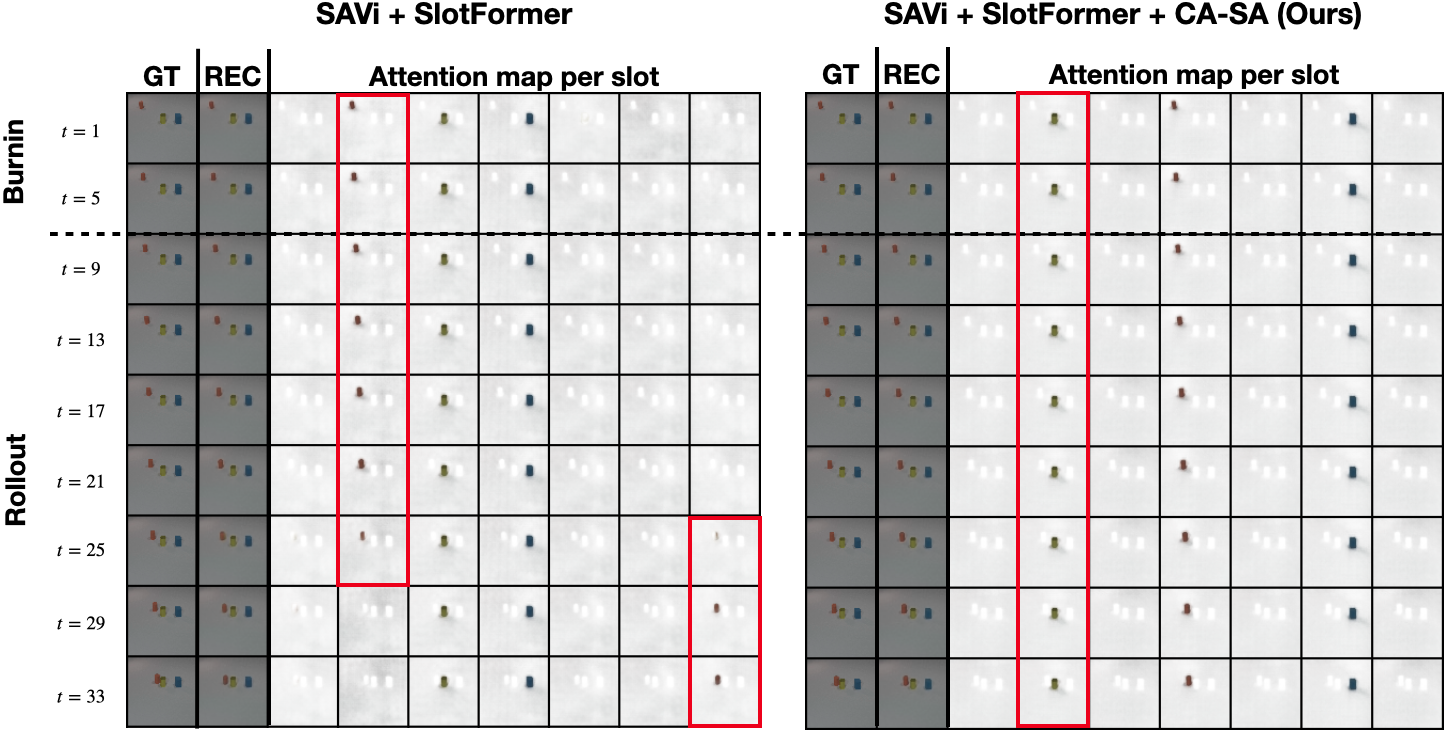}
    \caption{More generation results and predicted masks on CLEVRER. Red square indicate slots which temporal consistency is improved by adding \proposedshort.}
    \label{fig:more_rollout_clevrer}
\end{figure}

\begin{figure*}[t]
    \centering
    \includegraphics[width=\textwidth]{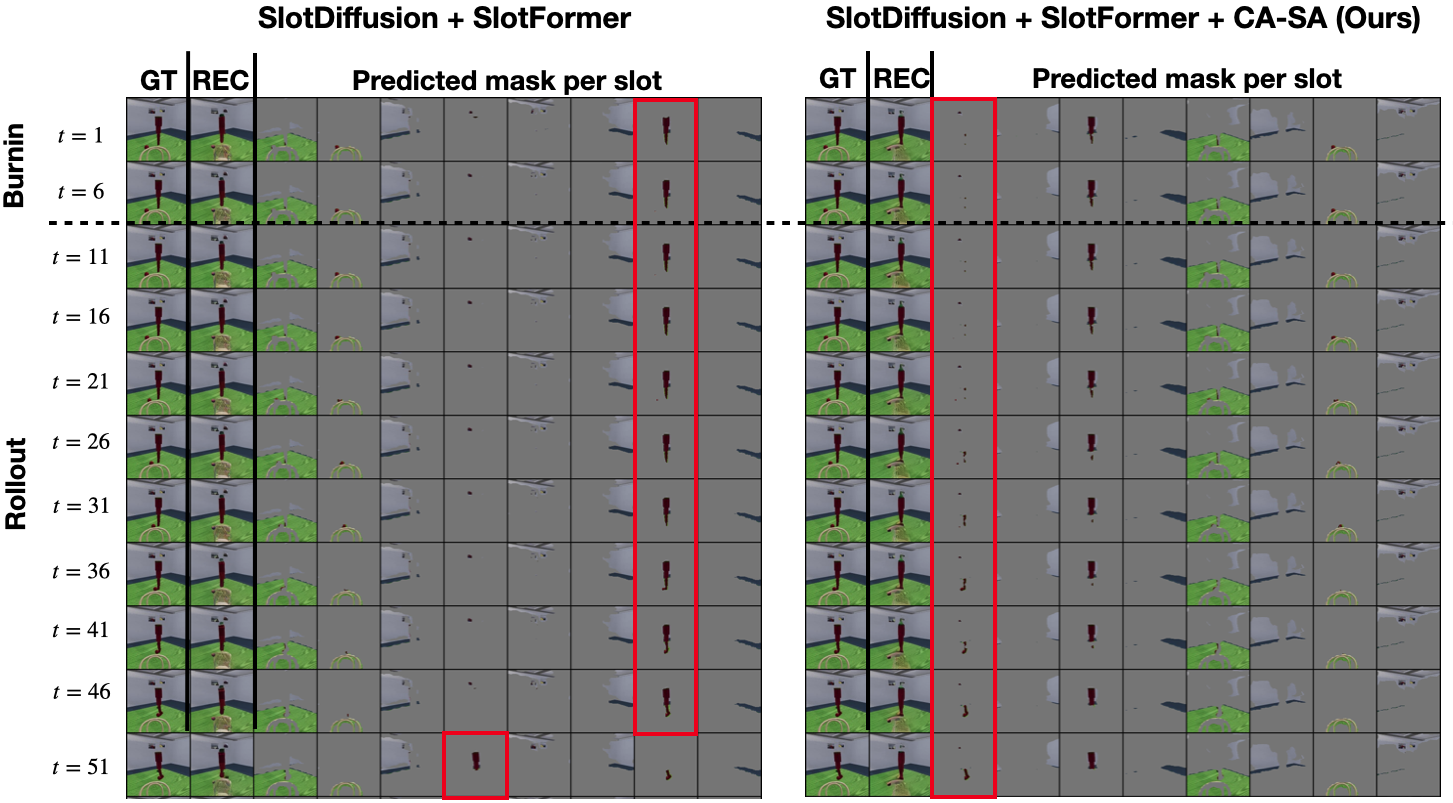}
    \includegraphics[width=\textwidth]{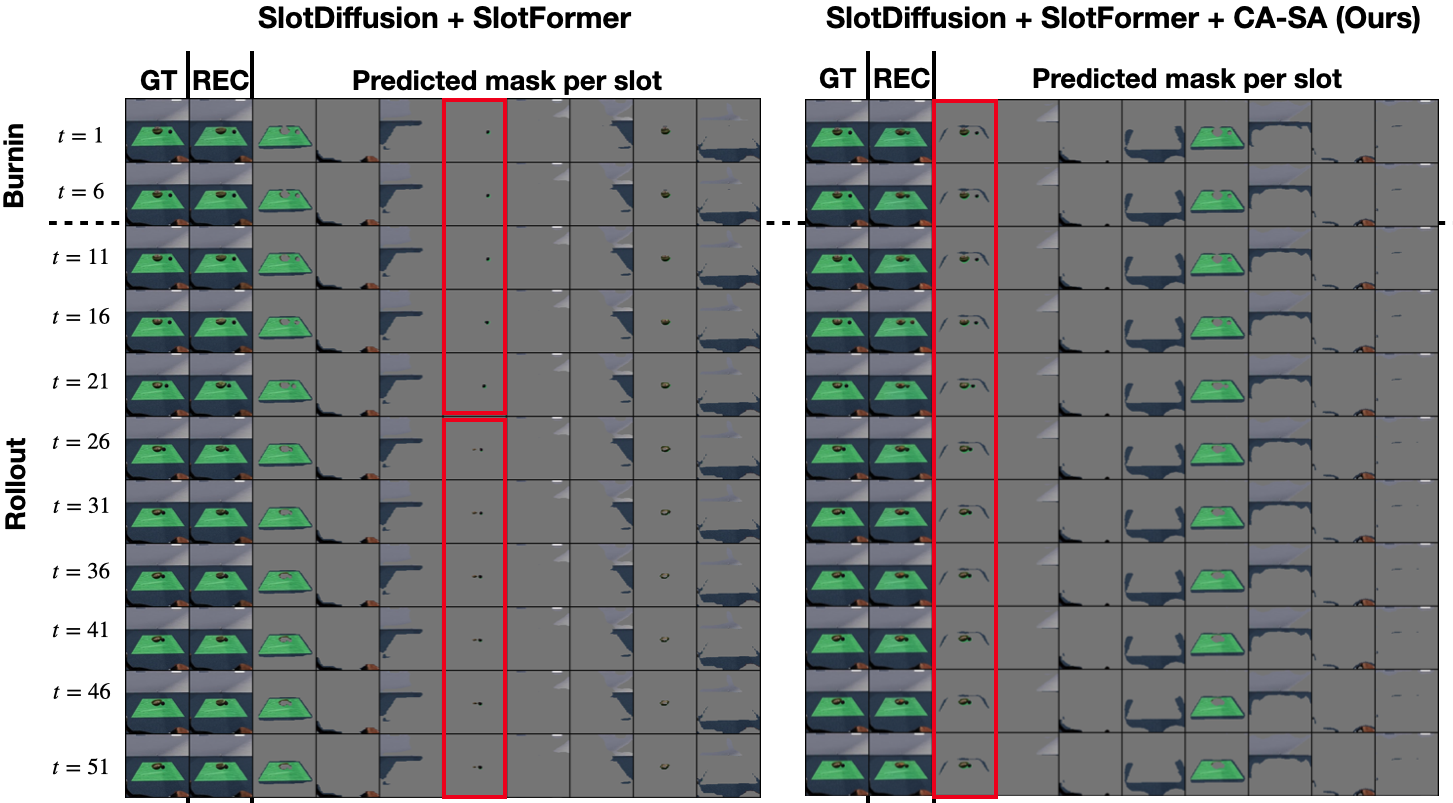}
    \caption{More generation results and predicted masks on Physion. Red square indicate slots which temporal consistency is improved by adding \proposedshort.}
    \label{fig:more_rollout_physion}
\end{figure*}

We provide further qualitative results of generated results and predicted attention maps on CLEVRER and Physion datasets in \autoref{fig:more_rollout_clevrer} and \autoref{fig:more_rollout_physion}, respectively.

\section{Additional Results on VQA Task} \label{appendix:additional_vqa}
We provide the accuracy per type of questions of CLEVRER in \autoref{table:vqa_clevrer_detailed}. We report the per-scenario accuracy on Physion for the model trained with rollouts in \autoref{table:vqa_physion_detailed}.

\begin{table*}[h]
  \caption{Detailed evaluation of video question answering task on CLEVRER dataset.}
  \label{table:vqa_clevrer_detailed}
  \centering
  \resizebox{\textwidth}{!}{%
  \begin{tabular}{cccccccc}
    \toprule
    \multirow{2}{*}{Model} & \multirow{2}{*}{Descriptive} & \multicolumn{2}{c}{Explanatory} & \multicolumn{2}{c}{Predictive} & \multicolumn{2}{c}{Counterfactual}\\
    & & per opt. (\%) & per ques. (\%) & per opt. (\%) & per ques. (\%) & per opt. (\%) & per ques. (\%)\\
    \midrule
    Aloe + SlotFormer$^*$ & 93.67 & 95.10 & 86.44 & 93.26 & 83.25 & 83.79 & 57.52 \\
    \midrule
    \textbf{Aloe + SlotFormer + \proposedshort (Ours)} & \textbf{94.10} & \textbf{96.56} & \textbf{90.65} & \textbf{94.85} & \textbf{90.28} & \textbf{86.65} & \textbf{64.47}
     \\
    \toprule
  \end{tabular}
  }
\end{table*}

\begin{table*}[h]
  \caption{Detailed evaluation of video question answering task on Physion dataset.}
  \label{table:vqa_physion_detailed}
  \centering
  \resizebox{\textwidth}{!}{%
  \begin{tabular}{ccccccccc|c}
    \toprule
    Model & Collide & Contain & Dominoes & Drape & Drop & Link & Roll & Support & Avg.\\
    \midrule
    SlotDiffusion + SlotFormer$^*$ & \textbf{75.3} & 63.3 & 49.2 & 51.3 & \textbf{65.3} & 59.3 & \textbf{68.0} & 70.0 & 63.9 \\
    \midrule
    \textbf{SlotDiffusion + SlotFormer + \proposedshort (Ours)} & 68.7 & \textbf{64.0} & \textbf{51.6} & \textbf{66.0} & 60.0 & \textbf{64.7} & 62.7 & \textbf{72.7} & \textbf{64.7} \\
    \toprule
  \end{tabular}
  }
\end{table*}

\end{document}